\documentclass[review,a4paper,number,sort&compress,3p]{elsarticle}
\usepackage{url,lineno,microtype,subcaption}
\usepackage[onehalfspacing]{setspace}
\usepackage{amsmath,mathtools}
\usepackage[hidelinks]{hyperref}
\usepackage{siunitx}
\usepackage{amssymb}
\DeclareMathOperator*{\argmin}{arg\,min}

\newcommand{\Cm}{C_\mathrm{m}}
\newcommand{\Vm}{V_\mathrm{m}}
\newcommand{\Iion}{I_\mathrm{ion}}
\newcommand{\Istim}{I_\mathrm{stim}}
\newcommand{\Imax}{I_\mathrm{max}}
\newcommand{\vx}{\mathbf{x}}
\newcommand{\vw}{\mathbf{w}}
\newcommand{\tGm}{\mathbf{G}_\mathrm{m}}
\DeclareMathOperator{\divg}{div}

\def\vec   #1{\mbox{\boldmath $#1$}{}}
\def\scas  #1{\mbox{{\scriptsize{${\rm{#1}}$}}}{}}

\def\ten   #1{\mbox{\boldmath $#1$}{}}

\def\bltr  #1{\mbox{\sffamily{\bfseries{{#1}}}}}

\journal{arXiv}

\begin{document}
\onecolumn

\begin{frontmatter}
%%%%%%%%%%%%%%%%%%%%%%%%%%%%%%%%%%%%%%%%%%%%%%%%%%%%%%%%%%%%%%%%%%%%%%%%
\title{Fast characterization of inducible regions of atrial fibrillation models with multi-fidelity Gaussian process classification}

\author[CCMC]{Lia Gander}
\author[CCMC]{Simone Pezzuto}
\author[CCMC]{Ali Gharaviri}
\author[CCMC]{Rolf Krause}
\author[PENN]{Paris Perdikaris}
\author[01,04,05]{Francisco~Sahli Costabal\corref{cor1}}
\cortext[cor1]{Corresponding author}
\ead{fsc@ing.puc.cl}

\address[CCMC]{Center for Computational Medicine in Cardiology,
Euler Institute,
Universit\`a della Svizzera italiana, Lugano, Switzerland}

\address[PENN]{Department of Mechanical Engineering and Applied Mechanics, University of Pennsylvania, Philadelphia, Pennsylvania, USA}

\address[01]{Department of Mechanical and Metallurgical Engineering, School of Engineering, Pontificia Universidad Cat\'olica de Chile, Santiago, Chile}

\address[04]{Institute for Biological and Medical Engineering, Schools of Engineering, Medicine and Biological Sciences, Pontificia Universidad Cat\'olica de Chile, Santiago, Chile}

\address[05]{Millennium Nucleus for Cardiovascular Magnetic Resonance}

\begin{abstract}

Computational models of atrial fibrillation have successfully been used to predict optimal ablation sites. A critical step to assess the effect of an ablation pattern is to pace the model from different, potentially random, locations to determine whether arrhythmias can be induced in the atria. In this work, we propose to use multi-fidelity Gaussian process classification on Riemannian manifolds to efficiently determine the regions in the atria where arrhythmias are inducible. We build a probabilistic classifier that operates directly on the atrial surface.  We take advantage of lower resolution models to explore the atrial surface and combine seamlessly with high-resolution models to identify regions of inducibility. When trained with 40 samples, our multi-fidelity classifier shows a balanced accuracy that is 10\% higher than a nearest neighbor classifier used as a baseline atrial fibrillation model, and 9\% higher in presence of atrial fibrillation with ablations. We hope that this new technique will allow faster and more precise clinical applications of computational models for atrial fibrillation. All data and code accompanying this manuscript will be made publicly available at: \url{https://github.com/PredictiveIntelligenceLab/JAX-BO}.

\begin{keyword}
Machine learning, Cardiac electrophysiology, Atrial fibrillation, Gaussian processes, Riemannian manifolds, Active learning
\end{keyword}  %All article types: you may provide up to 8 keywords; at least 5 are mandatory.
\end{abstract}

\end{frontmatter}

\section{Introduction}

\begin{figure}[t]
\includegraphics[width=\textwidth]{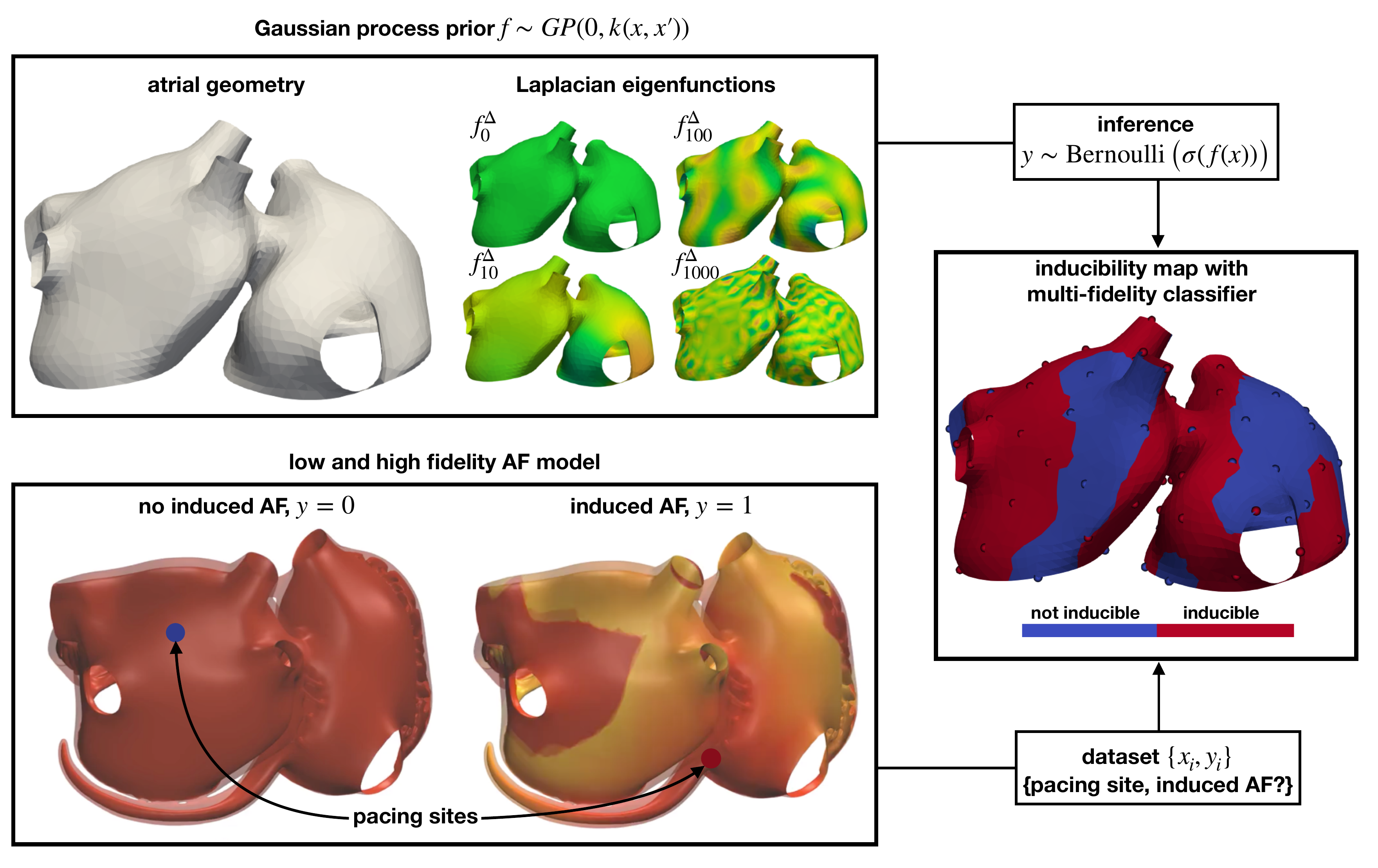}
\caption{We predict the regions where AF can be induced using a multi-fidelity Gaussian process classifier. We use the Laplacian eigenfunctions of the atrial geometry to efficiently construct a Gaussian process covariance function that operates directly on the manifold surface. We pace different sites in computational models of AF with low and high resolution to create a dataset to train our classifier. In the end, we obtain an inducibility map that can be used to assess treatments.}
\label{fig01}
\end{figure}

Atrial Fibrillation (AF) is the most common cardiac arrhythmia and a significant contributor to morbidity and mortality~\cite{virani2021heart}.
AF is characterized by a chaotic electrical activity of the atria and perpetuated by multiple re-entrant wavelets propagating in the atrial tissue. It has been shown in several studies that in patients in the early stages of AF (paroxysmal AF), the chaotic activity is originated mainly from the pulmonary veins (PVs)~\cite{haissaguerre1998,chen1999aha}. Thus, PV isolation (PVI) is the cornerstone of AF treatment at this point~\cite{Kawai2019}. Here, ablation lines around the PVs are created to electrically isolate them. However, in patients with a persistent form of AF, PVI efficacy remains sub-optimal~\cite{Kawai2019,verma2015approaches}. The detriment in the effect of this treatment in persistent AF patients is caused mainly by the shift of electrical abnormalities in the PVs to other locations and higher degrees of structural remodelling~\cite{Kawai2019,boyle2019}. Targeting arrhythmic substrates in persistent AF patients, in addition to PVI, could not demonstrate any benefit, as these treatment approaches do not incorporate strategies to find optimal ablation targets according to the AF mechanism~\cite{verma2015approaches}. Furthermore, the high inter-individual variability in fibrosis distributions~\cite{McDowell2015,boyle2019} and sources maintaining AF indicates an urgent need for patient-specific approaches. 

Simulations, conducted in computational atrial models, have recently been used to develop mechanistic insights into the perpetuation and ablation of persistent AF patients with atrial fibrosis~\cite{McDowell2015,boyle2019,Roney2019,loewe2019patient}.
A common approach to investigate AF is to stimulate a high-fidelity model from different pacing sites and observe whether this arrhythmia was induced or not. With these simulations, it is possible to create an inducibility map that shows the regions of the atria where AF will manifest if stimulated. Moreover, this map can be reduced into one metric, the inducibility, which corresponds to the fraction of the tissue where AF can be induced. This quantity is useful to compare different ablation treatments, as the most efficient intervention will be the one that reaches the lowest inducibility with the lowest amount of ablation. However, creating inducibility maps using these high-resolution models is computationally expensive. The complete exploration of all the potential sites that could trigger an arrhythmia is currently unfeasible~\cite{loewe2019patient}. For this reason, a number of alternatives have been proposed. A viable option is to design a pacing protocol that maximizes the chance of inducing AF~\cite{azzolin2021}. Alternatively,
the computational cost per simulation could be reduced by a faster implementation of the AF model, e.g., based on GPGPU~\cite{kaboudian2019}. Additionally, low fidelity models provide an approximation that could be based on simplified physics, e.g., Eikonal models~\cite{fu2013fast,quaglino2018fast}, reduced-order modeling~\cite{fresca2020deep,pagani2021enabling} or simply on a coarser discretization~\cite{Quaglino2019MFMC}.
Low fidelity models are faster, but imprecise and may not accurately reproduce the inducibility regions of the high fidelity model.
However, since the low fidelity models are constructed from the high fidelity model, a certain degree of statistical correlation of quantities of interest is to be expected.
Multi-fidelity approaches can exploit this inter-model correlation to improve the accuracy of the estimators for a fixed total cost or, equivalently, to reduce the total cost of estimation for a targeted accuracy~\cite{perdikaris2016multifidelity,sahli2019MF}.
This is achieved by offsetting most of the computational burden to the low-fidelity model.
Moreover, the overall computational cost could also be further reduced by carefully selecting the training points.
To this end, Bayesian decision making strategies, commonly referred to as active learning \cite{cohn1996active}, can provide a principled way for judiciously selecting new observations towards improving classification accuracy.
The process consists in adding points iteratively in the locations where the uncertainty is greater~\cite{kapoor2007active,Gramacy2017,sahli2020classifying}.

The problem of creating an inducibility map can be seen as a classification problem, from a machine learning perspective. The labels, in this case, are the occurrence or absence of AF when we pace the model from a specific site, which corresponds to the input. Although this may seem a trivial task, for which many tools are available, it is not straightforward when the classification domain is a Riemannian manifold, such as the atrial surface. In this case, points that may be close in the Euclidean space might be apart in the manifold due to its topology. There has been recent attention in the machine learning community on formulating effective Gaussian process (GP) models for supervised learning on Riemannian manifolds \cite{maternGP}. Gaussian processes tend to perform well when the amount of data available is limited, and, due to their Bayesian nature, they provide built-in uncertainty in the predictions. However, current approaches, also adopted in the cardiac modeling community, have focused on the regression case \cite{Coveney2019, coveney2020}. Performing classification with Gaussian processes is a challenging task, as there is no closed expression of likelihood and requires different types of approximations to perform statistical inference \cite{rasmussen2006gaussian}. In this work, we develop Gaussian process classifiers that can operate on manifolds, such as the atrial surface. We extend this tool to seamlessly combine different levels of data fidelity by creating a multi-fidelity Gaussian process classifier. In the specific context of AF, we aim to develop a method that allows us to comprehensively determine atrial regions, for a specific structural remodeling pattern, that, if stimulated could successfully initiate AF, creating an inducibility map \emph{in-silico}. In particular, our low fidelity model is based on a coarser spatial discretization of the atrial geometry and on a larger time step in the solution of the electrophysiology equations. The inducibility map is reconstructed using a multi-fidelity Gaussian process classifier, resulting in a function on the atrial surface taking boolean values, depending on whether AF is or is not inducible when pacing from a given location. We will demonstrate that this approach is more efficient and accurate than other classifiers, and even single-fidelity methods, for cases with and without ablation treatments.

% {\color{red}This is a good place to discuss GP on manifolds.
% FSC: this paper does GP classification on certain types of manifolds, but it is not general: \cite{lin2019extrinsic}.}

The manuscript is organized as follows. In Section~\ref{sec:methods} we present the AF model and the classification method,
obtained by extending the classic Gaussian process classification on manifolds.
We also present the multi-fidelity approach, as well as the active learning scheme employed to sequentially acquire new information.
Section~\ref{sec:experiments} is devoted to the numerical experiments.
Specifically, we propose a numerical assessment of the classifiers, including two case studies involving the characterization of inducibility regions of atrial models.
The discussion in Section~\ref{sec:discussion} concludes the manuscript.

\section{Methods}
\label{sec:methods}

\subsection{Atrial modeling}
\label{sec:modeling}

\begin{figure}[htb]  
\centering
\includegraphics[width=\textwidth,keepaspectratio]{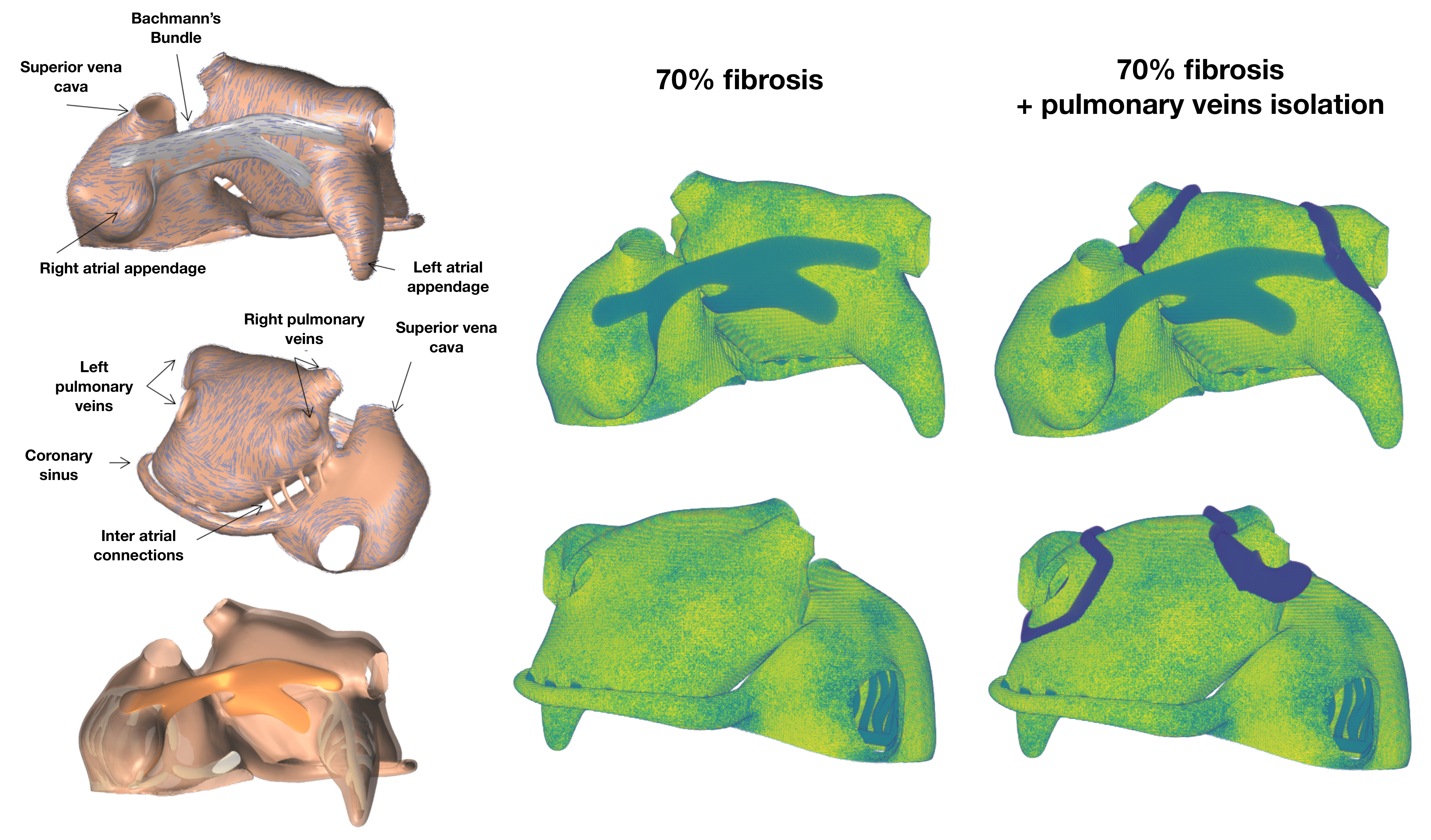}
\caption{\textbf{Anatomy of the computer model of AF.} Left: anatomical model of the atria, used for simulations, and three layers of fiber orientations. Top left, anterior view of the model and epicardial layer of fiber orientations, indicated by blue lines. Middle left, posterior view of the model, with the epicardial layer of fiber orientations. Bottom left, endocardial bundles in both atria (anterior view).
Middle column, atrial model with 70\% fibrosis (in yellow). Right column, atrial model with 70\% fibrosis and pulmonary veins isolation (in blue).}
\label{fig:anatomy}
\end{figure}

In this work, we consider a well-established model of atrial fibrillation (AF), previously
described~\cite{gharaviri2020epicardial}.
The model is based on an accurate anatomical description of the human atria and the whole torso, obtained from MRI data. The atrial anatomy, depicted in Figure~\ref{fig:anatomy}, contains several key anatomical features, shown to play an essential role in AF initiation, including atrial wall thickness heterogeneity~\cite{wang1995architecture, ho1999anatomy}, endocardial bundles~\cite{ho2002atrial, potse2016p}, Bachmann's bundle, interatrial bundles, crista terminalis; left and right atrial appendages trabecular network. Moreover, the model also contains coronary sinus (CS) with a few fibers connecting it to the left posterior wall~\cite{chauvin2000anatomic}.
%\textcolor{red}{FSC: is there a word missing here?}.
Three layers of fiber orientation were incorporated in the atrial model~\cite{ho2002atrial,ho2009importance}.
%Finally, the atria are embedded in an inhomogeneous torso model which comprises the whole heart, lungs, esophagus, major blood masses,
%skeletal muscles. Several electrodes are placed on the chest in order to simulate body surface potential maps in most common configurations.

The electrical activity is modeled with the monodomain system, which
reads as follows
\begin{equation}\label{eq:mono}
\begin{cases}
\chi \bigl( \Cm \partial_t \Vm + \Iion(\Vm,\vw) + \Istim(\vx,t) \bigr) = \divg(\tGm\nabla\Vm), & 
(\vx,t)\in\Omega\times(0,T], \\
\partial_t \vw = \mathbf{g}(\Vm, \vw), & (\vx,t)\in\Omega\times(0,T], \\
\tGm\nabla\Vm\cdot \mathbf{n} = 0, & (\vx,t)\in\partial\Omega\times(0,T], \\
\Vm(\vx,0) = V_0, \quad \vw(\vx,0) = \vw_0, & \vx\in\Omega,
\end{cases}
\end{equation}
where $\Vm(\vx,t)$ is the transmembrane potential, $\vw(\vx,t)$ is a vector of ion gating and concentration
variables, $\Omega$ is a domain describing the active myocardium, 
$\Cm = \SI{1}{\micro\farad\per\square\cm}$ is the membrane capacitance,
$\chi = \SI{800}{\per\mm}$ is the membrane surface-to-volume ratio,
$\Istim$ is the current stimulus, $\tGm(\vx)$ is the monodomain conductivity tensor,
and $\Iion$ and $\mathbf{g}$ describe the ionic model.  In particular, we consider
the Courtermanche-Ramirez-Nattel model \cite{courtemanche1998ionic} adapted to an AF
phenotype~\cite{gharaviri2017disruption, gharaviri2020epicardial}, with additional adaptations
to guarantee numerical stability \cite{potse2019inducibility}.
The initial condition $(V_0,\vw_0)$ corresponds to the resting state.

The conductivity tensor $\tGm$ is defined as the harmonic average of intra- and extra-cellular conductivity
tensors, both assumed transversely isotropic with higher electric conductivity longitudinally
to the local fiber direction.  In particular, intracellular longitudinal and cross conductivity are
respectively set \SI{3}{\milli\siemens\per\cm} and \SI{0.3}{\milli\siemens\per\cm}, while the
extracellular conductivities are \SI{3}{\milli\siemens\per\cm} and \SI{1.2}{\milli\siemens\per\cm}, respectively.
In the Bachmann's bundle, faster conduction is obtained with a longitudinal intracellular conductivity
of \SI{9}{\milli\siemens\per\cm}. Finally, the region between the superior and inferior vena
cava is assumed isotropic, with all conductivities set to \SI{1.5}{\milli\siemens\per\cm}.

In the numerical experiments for this study, we consider two atrial models. Both include severe fibrosis, as typical for persistent AF~\cite{gharaviri2020epicardial}. The second model,
however, also implements a standard-of-care PVI, achieved by isolating the tissue
surrounding the pulmonary veins with a line ablation. The ablation is performed by removing the
offended portion of the tissue from the domain $\Omega$ (Figure~\ref{fig:anatomy}).  Regarding the fibrosis modeling,
we consider endomysial fibrosis, which is modeled by formally imposing no cross
intracellular conductivity in fibrotic regions. Such regions are unevenly distributed across the tissue
so to create islands or patches showing a higher degree of fibrosis \cite{jakes2019perlin} (see Figure~\ref{fig:anatomy}),
as previously described~\cite{pezzuto2019sampling,gharaviri2020epicardial}. The total proportion
of fibrotic tissue is 70\% in all the numerical experiments.

The numerical solution of Eq.~\eqref{eq:mono} is based on a second-order finite difference scheme for the
spatial discretization, and a fully explicit first-order Euler scheme for time stepping~\cite{potse2006comparison}.
The Rush-Larsen scheme is adopted to update the gating variables. The computational domain is discretized using a uniform mesh
with hexahedral elements of side $h$.
For the high-fidelity simulations, we considered a fine mesh with $h=\SI{0.2}{\mm}$ and a time step of $\Delta t= \SI{0.01}{\ms}$.
For the low-fidelity simulations, we double the discretization parameters, with $h=\SI{0.4}{\mm}$ and $\Delta t = \SI{0.02}{\ms}$.
The coarsening of the grid is performed by employing a majority rule to determine the tissue type and fiber orientation of the coarse hexahedral elements
from the eight sub-elements of the fine mesh. Moreover, 
the coarse model assumes a reduced surface-to-volume ratio $\chi = \SI{450}{\per\cm}$ to balance out the expected reduction in conduction velocity
due to a coarser space discretization~\cite{pezzuto2016space}.
All simulations are performed with the Propag-5 software~\cite{potse2006comparison, krause2012hybrid} on the
Swiss National Supercomputing Centre (CSCS).
For one simulation with $T=\SI{4}{\s}$, the compute time of the high-fidelity model is $\SI{1}{\hour}\SI{40}{\minute}$ with 8 nodes, whereas the compute time of the low-fidelity model is $\SI{14}{\minute}$ with 4 nodes. This means that the low-fidelity model is approximately 16 times faster than the high-fidelity model.

\subsection{Pacing protocol for atrial fibrillation}
\label{sec:pacing}

\begin{figure}[h!]  
\includegraphics[width=\textwidth]{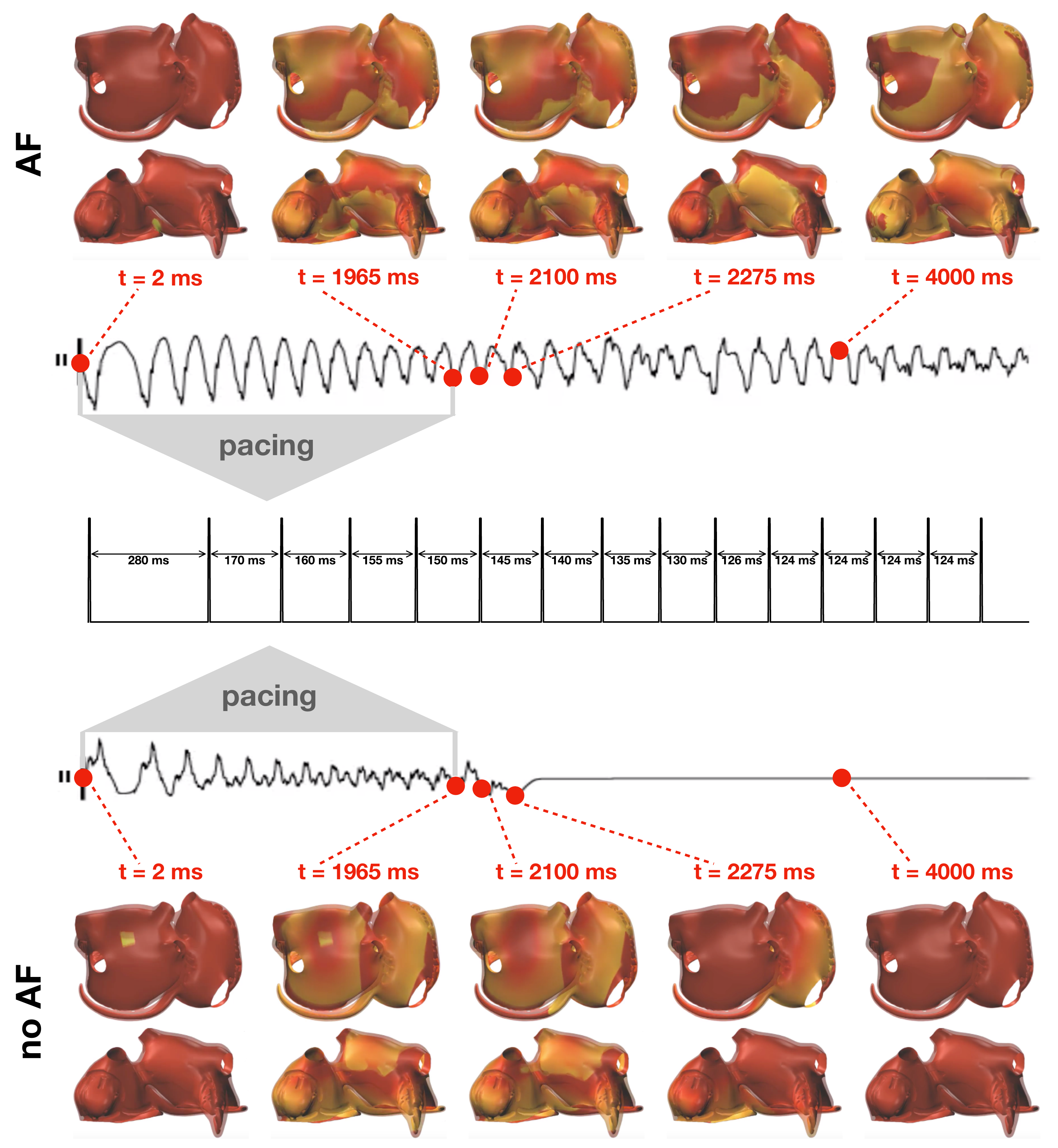}
\caption{\textbf{Inducibility of AF in the computer model.} Two simulations with different inducibility outputs. Stimulating from the top pacing site results in self-sustained activity and AF induction. Instead, after the stimulation from the bottom pacing site, the potential wave vanishes and there is no AF induction. The middle plot illustrates the pacing protocol.}
\label{fig:AFexamples}
\end{figure}

The stimulation protocol, encoded in the function $\Istim(\vx,t)$, is defined by
a point $\vx_\mathrm{stim}\in\Omega$ and a vector of distinct times
$\boldsymbol{\tau}_\mathrm{stim} = \bigl\{ \tau_j \bigr\}_{j=1}^{N_\mathrm{stim}}$
through the expression
\begin{equation}
\label{eq:stim}
\Istim(\vx,t;\vx_\mathrm{stim},\boldsymbol{\tau}_\mathrm{stim}) = \begin{cases} \Imax,  &
(\vx,t) \in B_r(\vx_\mathrm{stim}) \times \bigcup_{j=1}^{N_\mathrm{stim}} [\tau_j,\tau_j+\delta], \\
0, & \text{otherwise}, \end{cases}
\end{equation}
where $B_r(\vx_\mathrm{stim}) = \{ \vx \in \Omega: \vx_\mathrm{stim} \le \vx \le \vx_\mathrm{stim} + r \}$
is a $r$-neighborhood of the stimulation site, and $\delta > 0$ is the stimulus duration.  In this study,
the vector $\boldsymbol{\tau}_\mathrm{stim}$ is fixed as in Figure~\ref{fig:AFexamples}, which consists
in a series of $N_\mathrm{stim} = 14$ stimuli with decreasing temporal distance.  Each stimulus lasts
$\delta = \SI{4}{\ms}$, and has a strength $\Imax = \SI{800}{\micro\A\per\square\cm}$ with
a fixed radius $r = \SI{0.8}{\cm}$. The size and strength of the stimuli are sufficiently large to maximize the chance of inducing AF.

The objective of this work is to determine a set $\mathcal{A}\subset\Omega$ such that
pacing from $\vx_\mathrm{stim}\in\mathcal{A}$ leads to a sustained episode of AF.
In particular, we are interested in approximating the indicator function of $\mathcal{A}$,
denoted by $\mathbf{F}: \Omega\to \{ 0, 1 \}$m such that $\mathbf{F}^{-1}(1) = \mathcal{A}$.
The induction of AF is not successful when $(\Vm,\vw)$ asymptotically approach the
resting state after the last stimulus has been delivered. Otherwise, if a self-sustained
activity is still present at the end of the simulation, the induction of AF is successful.
The idea is summarized in Figure~\ref{fig:AFexamples}.
For sake of simplicity, in this work there is no distinction between a truly AF episode
and an atrial flutter, which could be understood as a periodic solution of the
monodomain system.

The overall inducibility, which reflects the fraction of the tissue where AF can be initiated, follows immediately from the definition of
$\mathcal{A}$ as
\[
\mathcal{I} = \frac{|\mathcal{A}|}{|\Omega|} = \frac{1}{|\Omega|} \int_{\Omega} F(\vx) \: \mathrm{d}\vx.
\]
Interestingly, the formula generalizes to the case of non-uniformly distributed
ectopic foci. Let $\rho(\vx)$ be the probability density function of the distribution
of foci, then the inducibility can be obtained as
\[
\mathcal{I}_\rho = \int_{\Omega} F(\vx) \rho(\vx) \: \mathrm{d}\vx.
\]
In this way, for instance, it is possible to account for a higher density of
ectopic activity around the pulmonary veins and is fibrotic regions. In this
work, we will only consider a uniform distribution of foci, equivalent to select
$\rho(\vx) = |\Omega|^{-1}$.

\subsection{Classification with Gaussian processes}
\label{sec:GPclass}

Next, we present the proposed methodology for learning the inducibility function $\mathbf{F}$
from a limited set of simulations.  We start by assuming that we have a data-set of $N$ input/output pairs $\mathcal{D} = \left\{ (\vx_i, y_i) \right\}^N_{i=1}$,
where $\vx_i \in \Omega$ and $y_i \in \{ 0, 1 \}$.  Since the atrial wall is thin,
we constraint the points to belong to a mid-wall smooth atrial surface $\mathcal{S}\subset\Omega$.
We remark however that there is no loss of generality in the following presentation,
as the methodology applies to the volumetric domain $\Omega$ in the same manner. Moreover,
since $y_i$ takes only binary values, we also restrict the scope of this work to binary classification.
We also note that it is straightforward to extend this framework to the multi-class classification setting.

% $= \left\{\ten{X}, \vec{y}\right\}$.
% The input $\ten{X}$ represents a point in the manifold for $N$ training examples and the outputs $\vec{y}$ contain the $N$ discrete class labels for the corresponding inputs. Here, we restrict the scope of this work to binary classification, thus the labels can only take on two values $y_i = \{0,1\}$. We note that it is possible to extend this framework to the multi-class classification setting.
The classical formulation of Gaussian process classification defines an intermediate variable
which is computed from a latent function $f(\vx)$~\cite{rasmussen2006gaussian}. Throughout this paper,
we will assume standardized data-sets and work with zero-mean Gaussian process priors of the form
$f \sim \mathcal{GP}\,(\vec{0},k(\vx,\vx'; \theta))$. Here, $k(\cdot,\cdot;\theta)$ is a covariance kernel function,
which depends on a set of parameters $\theta$. We adopt a fully Bayesian treatment and prescribe prior distributions 
over these parameters, which we will specify later~\cite{neal1999regression}. To obtain class probability predictions
we pass the Gaussian process output $f$ through a nonlinear warping function $\sigma: \mathbb{R}\to[0,1]$, such that the output 
is constrained to $[0,1]$, rendering meaningful class probabilities. 
We define the conditional class probability as $\pi(\vx) = \mathbb{P}[y =  1 | \vx] = \sigma(f(\vx))$. A common choice for $\sigma(f)$ is
the logistic sigmoid function $\sigma(f) = (1 + \exp{(-f)})^{-1}$, which we will use throughout this work.
We assume that the class labels are distributed according to a Bernoulli likelihood with probability $\sigma(y)$~\cite{Nickisch2008}.

\subsection{Gaussian process priors on manifolds}
\label{sec:manifolds}

A crucial step in building a Gaussian process classifier is the choice of the kernel function. A popular choice is the Mat\'ern kernel, which explicitly allows one to encode smoothness assumptions for the latent functions $f(\vx)$~\cite{rasmussen2006gaussian}.
In a Euclidean space setting, the kernel function has the form \cite{rasmussen2006gaussian}:
\begin{equation}
k(\vx,\vx', \theta) = \eta^2 \frac{2^{1-\nu}}{\Gamma(\nu)}\left(\sqrt{2\nu} \frac{\|\vx - \vx'\|}{\ell}\right)^\nu K_{\nu}\left(\sqrt{2\nu} \frac{\|\vx - \vx'\|}{\ell}\right),
\label{eq:materneuc}
\end{equation}
where $\Gamma$ is the gamma function, and $K_{\nu}$ is the modified Bessel function of the second kind. The parameter $\eta$ controls the overall variance of the Gaussian process, the parameter $\ell$ controls the spatial correlation length-scale, and $\nu$ controls the regularity of the latent functions $f(\vx)$~\cite{rasmussen2006gaussian}. When $\nu \rightarrow \infty$, we recover the popular squared exponential kernel, also known as radial basis function, that yields a prior over smooth functions with infinitely many continuous derivatives.   

The form presented in Eq.~\eqref{eq:materneuc} is not suitable to be used on manifolds, as the atrial surface. A naive approach is to replace the Euclidean distance between points
with the geodesic distance on the manifold surface. Even though this approach may work for some cases, there is no guarantee that the resulting covariance matrix between input points will be positive semi-definite~\cite{maternGP}, a key requirement for a kernel function.  As a matter of fact, the choice of the kernel is problematic in this case. For instance, the Mat\'ern family does not yield positive definite kernels even on the sphere, except for a few exceptional choices of the parameters~\cite{gneiting2013}.  Here, we follow an alternative approach, implicitly
based on the solution of the following stochastic partial differential equation (SPDE)~\cite{whittle1963stochastic,lindgren2011}:
\begin{equation}\label{eq:spde}
\begin{cases}
(\kappa^2 \mathbf{I} - \Delta)^{\nu/2+d/4} u = \mathcal{W}, & \vx \in \Omega, \\
\mathbf{n}\cdot\nabla (\kappa^2 \mathbf{I} - \Delta)^j u = 0, & \mbox{$\vx \in \partial\Omega$, $j = 0, \ldots, \lfloor\frac{\nu-1}{2}+\frac{d}{4}\rfloor$},
\end{cases}
\end{equation}
where $-\Delta$ is the Laplace-Beltrami operator on the $d$-dimensional manifold, and $\mathcal{W}$ is the spatial Gaussian white noise on $\Omega$.
When $\Omega = \mathbb{R}^d$, the solution of the fractional SPDE is a Mat\'ern random field with $\kappa = \frac{\sqrt{2\nu}}{\ell}$~\cite{lindgren2011}.
However, compared to Eq.~\eqref{eq:materneuc}, the SPDE in Eq.~\eqref{eq:spde} trivially generalizes to manifolds with no loss of positive definiteness of the correlation kernel,
thanks to the properties of the pseudo-differential operator~\cite{maternGP}.  The correlation function can be explicitly written as follows.
Let $\{ (\lambda_i,\psi_i) \}_{i=0}^\infty$ be the eigenvalue/eigenfunction pairs of the Laplace-Beltrami operator with pure Neumann boundary conditions, that is
\begin{equation}\label{eq:eigen}
\begin{cases}
-\Delta \psi_i = \lambda_i \psi_i & \vx\in\Omega, \\
-\mathbf{n}\cdot\nabla \psi_i = 0, & \vx\in\partial\Omega,
\end{cases}
\end{equation}
for all $i\in\mathbb{N}$. Then, we can represent Matérn-like kernels on manifolds as
%Another method to create a random field in the manifold following a Mat\'ern kernel is to solve an stochastic partial differential equation \cite{whittle1963stochastic,pezzuto2019sampling}. In this case, the inverse of the pseudo-differential operator maps a Gaussian white noise on the manifold to a spatially-correlated random field with a specific correlation length and smoothness. However, in this case the kernel function is not directly available.  It is  To compute the kernel in the manifold, we follow a novel approach, which applies spectral theory to obtain the resulting kernels of the stochastic differential equation approach \cite{maternGP}. In summary, this methodology consists in two steps. First, we compute the Laplace-Beltrami operator $\Delta = \nabla \cdot \nabla$ of the manifold. Then, we obtain the eigenvalues $\lambda_{\rm i}$ and eigenfunctions $f_{\rm i}(x)$. Now, we can approximate the Mat\'ern kernel as follows:
\begin{equation}
k(\vx,\vx'; \theta) = \frac{\eta^2}{C}\sum_{\rm i = 0}^{\infty} \left(\frac{1}{\ell^2}+\lambda_i\right)^{-\nu - \frac{d}{2}}\psi_i(\vx)\psi_i(\vx')
\label{eq:materneig}
\end{equation}
where $C$ is a normalizing constant.  This eigen-decomposition also enables a direct solution of the SPDE, providing the following representation
of the Gaussian process prior:
\begin{equation}
f(\vx) \approx \frac{\eta^2}{C}\sum_{\rm i = 0}^{\infty} w_i\left(\frac{1}{\ell^2}+\lambda_{i}\right)^{-\frac{\nu}{2} - \frac{d}{4}}f_i(\vx),
\qquad w_i \sim \mathcal{N}(0,1)
\label{eq:maternprior}
\end{equation}
In practice, the eigendecomposition is truncated to a number $N_\mathrm{eig}$ of pairs.

In this work, we discretize the manifold $\mathcal{S} \in \mathbb{R}^3$ using a triangulated mesh and solve Eq.~\eqref{eq:eigen} using finite element shape functions.
As such, we can obtain the stiffness matrix $\ten{A}$ and mass matrix $\ten{M}$:
\begin{eqnarray}
\ten{A}_{ij} &=& \overset{n_{\scas{el}}}{\underset{\scas{e}=1}{\bltr{A}}} \int_{\mathcal{B}} \nabla N_{ i} \cdot \nabla N_{ j} d\mathcal{B}\\ 
\ten{M}_{ij} &=& \overset{n_{\scas{el}}}{\underset{\scas{e}=1}{\bltr{A}}} \int_{\mathcal{B}}  N_{i} N_{ j} d\mathcal{B},
\end{eqnarray}
where $\bltr{A}$ represents the assembly of the local element matrices, and $N$ are the finite element shape functions. 
Then, we solve the eigenvalue problem:
\begin{equation}
\ten{A}\vec{v} = \lambda \ten{M}\vec{v}
\end{equation}
In practice, to compute the kernel in Eq.~\eqref{eq:materneig} we use a portion of all the resulting eigenpairs, starting from the smallest eigenvalues. We also use the corresponding eigenvectors as the eigenfunctions with $f(\vec{x}_{\rm i}) = \vec{v}_{\rm i}$, where $\rm i$ is the node index at location $\vx_{\rm i}$. Given that the eigenvalue problem is solved only once as a pre-processing step, this methodology provides an efficient way to compute the kernel and the prior in a manifold.

\subsection{Bayesian inference}
\label{sec:inference}

 We finalize our Bayesian model description by prescribing the prior distributions for the kernel parameters. We assume the following distributions for the parameters $\theta = \{\eta, \ell\}$,
\begin{eqnarray}
  \eta & \sim & {\rm HalfNormal}(\sigma = 10000)\\
  \ell & \sim & {\rm Gamma}(\alpha = 1, \beta = 1).
\end{eqnarray}
The posterior distribution over the model parameters $\theta = \{\eta,\ell\}$ cannot be described analytically, and thus we must resort to approximate inference techniques to calibrate this Bayesian model on the available data. To this end, we use the NO-U-Turn sampler (NUTS) \cite{hoffman2014no}, which is a type of Hamiltonian Monte Carlo algorithm, as implemented in NumPyro \cite{phan2019composable}. We use one chain, and set the target accept probability to 0.9. The first 500 samples are used to adjust the step size of the sampler, and are later discarded. We use the subsequent 500 samples to statistically estimate the parameters $\theta$. 

Once we have completed the inference, we can make predictions $\vec{y}^*$ at new locations $\vec{x}^*$ in three steps. First, we compute the predictive posterior distribution of the latent function $f^*(\vec{x}^*) \sim \mathcal{N}(\mu(\vec{x}^*),  \Sigma(\vec{x}^{\ast}))$, which by construction follows a multivariate normal distribution, with a mean $\vec{\mu}$ and covariance $\Sigma$ obtained by conditioning on the available training data \cite{rasmussen2006gaussian}:
\begin{eqnarray}
  \mu(\vec{x}^{\ast})  & =  & k(\vec{x}^{\ast}, \vec{X}) \vec{K} ^{-1}\vec{f} 
\label{eq:posterior_mean} \\
  \Sigma(\vec{x}^{\ast}) & = & k(\vec{x}^{\ast}, \vec{x}^{\ast}) - k(\vec{x}^{\ast}, \vec{X}) \vec{K}^{-1} k(\vec{X},\vec{x}^{\ast}) \, ,
\label{eq:posterior_variance}
\end{eqnarray}
where the covariance matrix $\vec{K}\in\mathbb{R}^{N\times N}$ results from evaluating the kernel function $k(\cdot,\cdot;\theta)$ at the locations of the input training data $\vec{X}$ and $\vec{f} = f(\ten{X})$, respectively. We then proceed by sampling $\mu, \Sigma$ using model parameters drawn from the estimated posterior distributions of $\theta$ and $\vec{f}$. This will result in a number of random variables $\vec{f}^*$ that are independent and normally distributed, which we can be used to compute statistical averages as
\begin{eqnarray}
\hat{f}^* & \sim & \mathcal{N}(\hat{\mu}, \hat{\Sigma}) \\
\hat{\mu} & = & \frac{1}{N_{\rm s}}\sum_{\rm i = 1}^{N_{\rm s}} \mu_i \\
\hat{\Sigma} & = & \frac{1}{N_{\rm s}}\sum_{\rm i = 1}^{N_{\rm s}} \Sigma_i,
\end{eqnarray}
where $N_{\rm s}$ is the number of samples considered for $\theta$ and $\vec{f}$. We finally pass $\hat{f}^*$ through the logistic sigmoid function $\sigma$ to obtain a distribution of class probabilities $\vec{y}^*$.

\subsection{Multi-fidelity classification with Gaussian processes}
\label{sec:MFclass}
In this work, we will assume that we have 2 information sources of different fidelity. We will call the high fidelity, computationally expensive, and hard to acquire information source with the subscript $H$ and the inexpensive, faster to compute, low fidelity source with the subscript $L$. Now, our data set comes from these two sources $\mathcal{D} = \left\{(\vec{x}_{Li}, y_{Li})^{N_L}_{i=1}, (\vec{x}_{Hi}, y_{Hi})^{N_H}_{i=1} \right\} = \left\{\ten{X}, \vec{y}\right\}$. We will postulate two latent functions $f_H$ and $f_L$, respectively, that are related through an auto-regressive prior \cite{kennedy2000predicting}
\begin{equation}
    f_H(\vec{x}) = \rho f_L(\vec{x}) + \delta(\vec{x})
\end{equation}
Under this model structure, the high fidelity function is expressed as a combination of the low fidelity function scaled by $\rho$, corrected with another latent function $\delta(\vec{x})$ that explains the difference between the different levels of fidelity. Following \cite{kennedy2000predicting}, we assume Gaussian process priors on these latent functions
\begin{eqnarray}
f_L &\sim& \mathcal{GP}\,(\vec{0},k(\vec{x},\vec{x}'; \vec{\theta}_L)) \\
 \delta &\sim& \mathcal{GP}\,(\vec{0},k(\vec{x},\vec{x}'; \vec{\theta}_H)).
\end{eqnarray}
The vectors $\vec{\theta}_L$ and $\vec{\theta}_H$ contain the kernel hyper-parameters of this multi-fidelity Gaussian processes model. The choice of the auto-regressive model leads to a joint prior distribution over the latent functions that can be expressed as \cite{kennedy2000predicting}
\begin{equation}
\vec{f} =  \left[ 
  \begin{array}{c} \vec{f}_{L} \\ \vec{f}_{H} \end{array} \right] 
\sim \mathcal{N}\left(\left[\begin{array}{c} \vec{0} \\ \vec{0} \end{array} \right],  
\left[ \begin{array}{c c} \ten{K}_{LL} & \ten{K}_{LH}
 \\ \ten{K}_{LH} & \ten{K}_{HH}
  \end{array} 
  \right]\right), 
\end{equation}
with
\begin{equation}
\begin{array}{l@{\hspace*{0.1cm}}c@{\hspace*{0.1cm}}
              l@{\hspace*{0.1cm}}l@{\hspace*{0.1cm}}
              c@{\hspace*{0.1cm}}l@{\hspace*{0.1cm}}l}
  \ten{K}_{LL} 
& = & & k_{L}&(\vec{X}_{L},\vec{X}_{L}';\theta_{L})  \\
  \ten{K}_{LH} 
& = &\rho & k_{L}&(\vec{X}_{L},\vec{X}_{H}';\theta_{L}) \\
  \ten{K}_{HH} 
& = &\rho^2 & k_{L}&(\vec{X}_{H},\vec{X}_{H}';\theta_{L}) 
& + & k_{H}(\vec{X}_{H},\vec{X}_{H}';\theta_{H})  \,.
\label{eq:MF_K}
\end{array}
\end{equation}

The global covariance matrix $\ten{K}$ of this multi-fidelity Gaussian process model has a block structure corresponding to the different levels of fidelity, where $\ten{K}_{HH}$ and $\ten{K}_{LL}$ model the spatial correlation of the data observed in each fidelity level, and $\ten{K}_{LH}$ models the cross-correlation between the two levels of fidelity. We also have kernel parameters for the different levels of fidelity. We again use the Mat\'ern as described in Section~\ref{sec:manifolds},  which results in parameters $\theta_H = (\eta_H, \ell_H)$, and $\theta_L = (\eta_L, \ell_L)$. For these parameters and the scaling factor $\rho$, we consider the following prior distributions
\begin{eqnarray}
  \eta_H, \eta_L & \sim & {\rm HalfNormal}(\sigma = 10000)\\
  \ell_{H}, \ell_{L} & \sim & {\rm Gamma}(\alpha = 2, \beta = 2)\\
  \rho & \sim & {\rm Normal}(\mu = 0, \sigma = 10).
\end{eqnarray}

We can perform inference and prediction for this model in the same way as for the single fidelity classifier, as detailed in Section~\ref{sec:inference}. In particular, we can use equations~(\ref{eq:posterior_mean}) and (\ref{eq:posterior_variance}) with the entire covariance matrix $\ten{K}$ to obtain the conditional mean and covariance of $\vec{f}^*$.

\subsection{Active learning}
\label{sec:active}

Here, we take advantage of the uncertainty predictions that are inherent to Gaussian processes and are absent in other types of classifiers, such as nearest neighbor. Specifically, at each active learning iteration, we train the classifier, and select the next point that should be included in our training data-set by solving the following optimization problem \cite{kapoor2007active,sahli2019MF}:
\begin{equation}
    \vec{x}^{new} = \argmin_{\vec{x} \in \vec{X}_{cand}} \frac{|{\hat{\mu}(\vec{x})|}}{\sqrt{\hat{\Sigma}(\vec{x})}},
    \label{eq:AL}
\end{equation}   
where $\vec{X}_{cand}$ represents a set of candidate locations that can be acquired. In our case, we use all the nodes in the mesh as candidates, except the ones at the boundaries which have artificially high variance. This active learning criterion presents a good balance between exploitation (sampling near the classification decision boundary) and exploration (discovering new inducible regions). It can be seen as promoting the selection of points that tend to be located near the decision boundary ($\sigma(\mu = 0) = 0.5$), or points in regions with high uncertainty (as reflected by the posterior variance $\Sigma$). We keep adding points via this sequential active learning procedure until we have reached the desired number of samples.

\section{Numerical experiments}
\label{sec:experiments}

\subsection{Numerical assessment}
\label{sec:numerical}

We first create a synthetic example to test the performance of the proposed  classifier. In particular, we use a mesh that represents the left and right atria with 3298 nodes and 6335 triangles, respectively. We first normalize the coordinates of this geometry by centering it around the origin and then scaling it by the maximum standard deviation of its coordinates. Then, we generate Gaussian random fields on the atrial manifold with zero mean and the Mat\'ern covariance kernel, as detailed in Eq.~\eqref{eq:materneig}. We use 1000 eigenpairs to construct a computable kernel function approximation with $\nu = 3/2$ and $\eta = 1$. We consider different length scales to simulate inducibility regions and assess the performance of the classifier: $\ell = \{0.2,0.4,0.6,0.8,1.0\}$. Finally, we pass the resulting random field through the sigmoid function $\sigma$ to obtain values between zero and one, which we round to the nearest integer to create discrete labels.
\begin{figure}[h!]
\includegraphics[width=\textwidth]{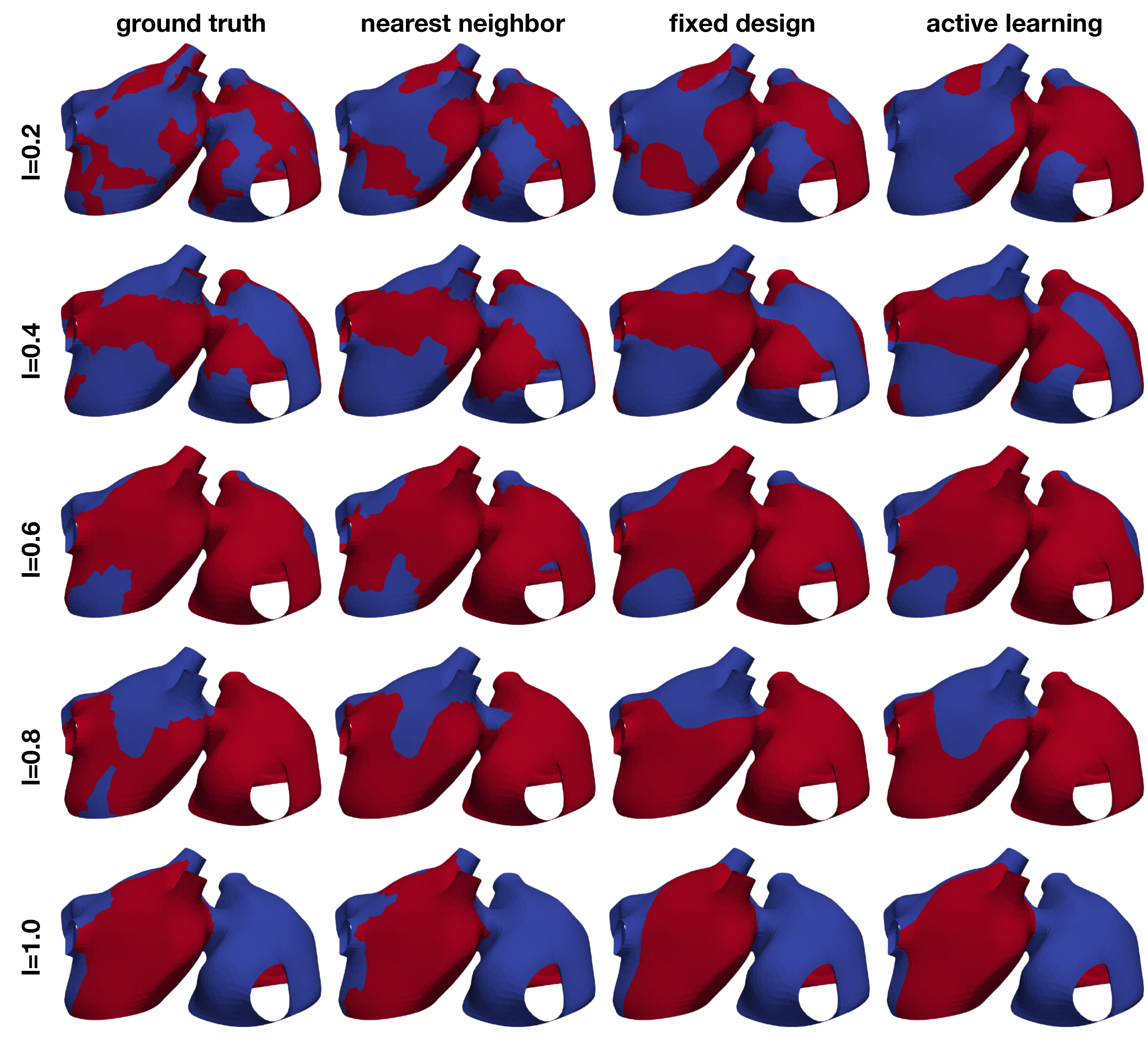}
\caption{\textbf{Numerical assessment of the Gaussian process classifier}. We create different random examples with different correlations lengths (first column) and train a nearest neighbor classifier (second column), a Gaussian process classifier trained with the same data-set as the nearest neighbor classifier (third column), and Gaussian process classifier that adaptively selects the training points through active learning (fourth column).}
\label{fig:NNvsFDvsAL}
\end{figure}

Examples of the resulting random fields can be seen in Figure~\ref{fig:NNvsFDvsAL}, left column. We compare three different classifiers. First, as a baseline benchmark, we create a nearest neighbor classifier. Here, the prediction of an unknown point is based on the label of the closest data point. Since we are working with a manifold, we use the geodesic distance to find the closest point, which we compute using the heat method \cite{Crane2013}. As a data-set, we use a fixed design spread through the manifold surface. To select the locations, we first randomly pick a node in the mesh, and then we add the node that is further away from the initial node using the geodesic distance. Then, we iterate, finding the point that is further away from all the nodes already included in the data-set, until we reached the desired data-set size. The second classifier that we consider is a Gaussian process classifier, as described in the previous sections, that is trained on the same fixed experimental design. The final classifier is also a Gaussian process classifier, which we train with the first 20 samples of the fixed experimental design, and then we apply the proposed active learning procedure.

In these examples we test the performance of the three different classifiers, using between 20 and 100 samples, and 10 different random fields for each of the 5 length scales selected. To take into account potential imbalances of classes in the examples generated, we use the balanced accuracy score. This metric is defined as the arithmetic mean of the sensitivity and specificity as
\begin{equation}
    \mbox{balanced accuracy} = \frac{1}{2}\left(\frac{\mbox{\# of predicted positives}}{\mbox{\# of real positives}}
                             + \frac{\mbox{\# of predicted negatives}}{\mbox{\# of real negatives}} \right).
\end{equation}
In contrast to conventional accuracy, this metric will reflect if a classifier is predominately predicting one class due to the higher proportion of samples present in the data-set.

\begin{figure}[h!]
\includegraphics[width=\textwidth]{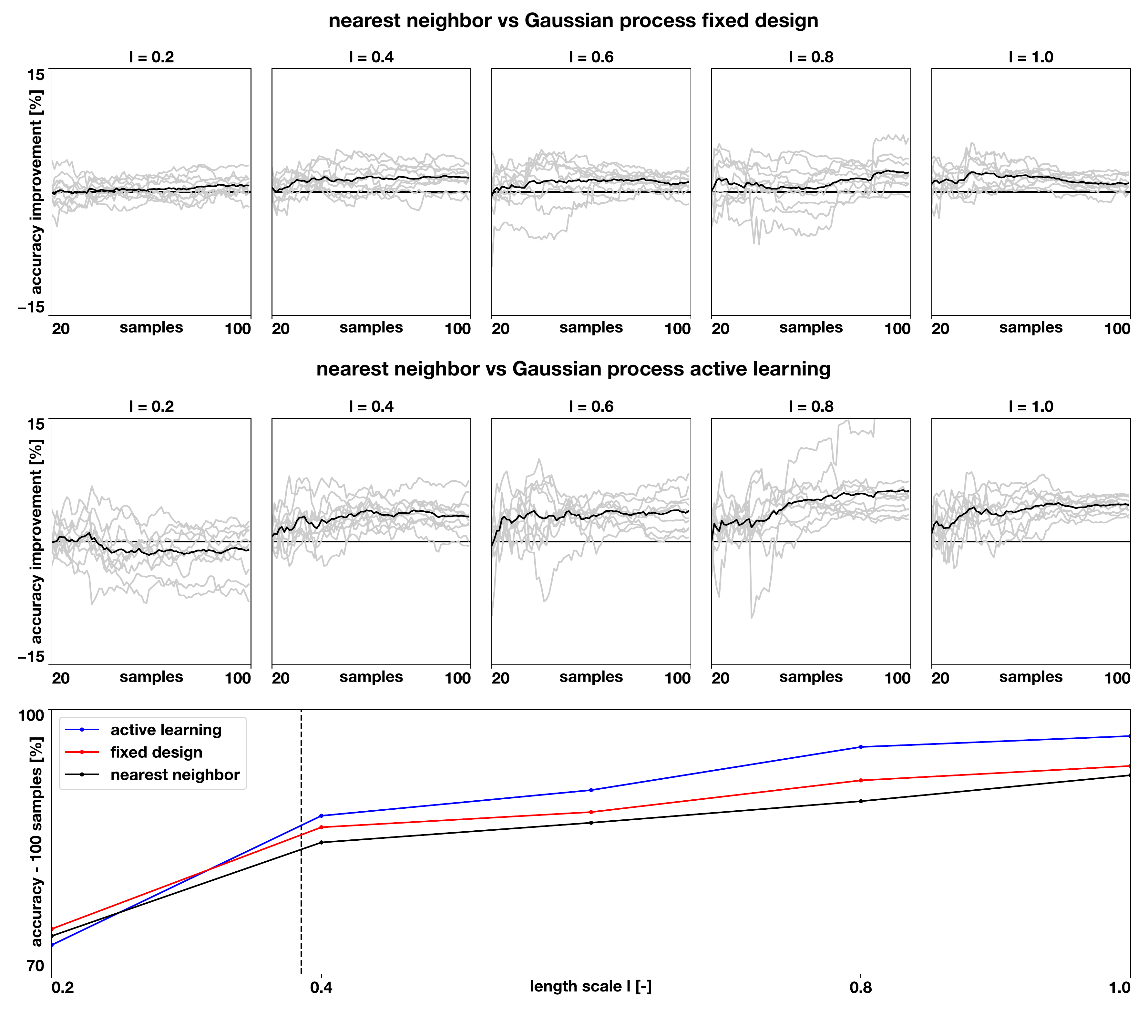}
\caption{\textbf{Accuracy of the numerical assessment.} We quantify the improvements in accuracy when using a Gaussian process the classifier versus the nearest neighbor classifier (top row) and when using a Gaussian process classifier with active learning, versus the baseline nearest neighbor classifier (middle row) for different length scales. The gray lines show the balanced accuracy improvements of the 10 examples for each length scale and the black line shows the mean improvement. The bottom row shows how the average balanced accuracy changes with length scale when the classifiers are trained with 100 samples. The dashed vertical line represents the average geodesic distance between training points of the fixed design.}
\label{fig:NNvsFDvsALplots}
\end{figure}

The results of this assessment are summarized in Figures~\ref{fig:NNvsFDvsAL} and~\ref{fig:NNvsFDvsALplots}. We first observe in Figure~\ref{fig:NNvsFDvsAL}, left column,  that the complexity of the classification regions increases as the correlation length scale is reduced. In the same Figure, we show the different classifiers trained with 100 samples. It is visually possible to note that the accuracy of the classifiers degrade as the length scale of the ground truth classification surface is decreased. For the length scale $\ell = 0.2$, some regions are not captured by the classifiers. We also note that the Gaussian process classification boundaries tend to be smoother than the nearest neighbor classifier. These differences are quantified in Figure~\ref{fig:NNvsFDvsALplots}. We first compare the improvements in accuracy between the nearest neighbor classifier and the Gaussian process classifier with a fixed design in the top row. These two methods are trained with identical data, and we observe that for most cases and number of samples, the Gaussian process classifier is more accurate than the nearest neighbor classifier. The accuracy improvements at 100 samples range on average from 0.8\% at $\ell = 0.2$ to 2.4\% at $\ell = 0.8$. Then, we compare the nearest neighbor classifier with the Gaussian process classier trained with active learning. These two classifiers only share the first 20 points of data. Then the active learning classifier judiciously selects the remaining samples attempting to maximize accuracy. We observe that the accuracy improvements are more pronounced with the active learning for $\ell = 0.4 - 1.0$. The average improvements at 100 samples range from 3.0\% at $\ell = 0.4$ to 6.2\% at $\ell = 0.8$. For $\ell = 0.2$, we see an average decrease in accuracy of 1.0\% at 100 samples. In the last row of Figure~\ref{fig:NNvsFDvsALplots} we see the average accuracies for the three classifiers at 100 samples, reflecting the improvements in accuracy obtained by the Gaussian process classifiers already described. We see that all classifiers tend to decrease their accuracy as the length scale decreases, which coincides with the increased complexity of classification boundaries for lower length scales seen in Figure~\ref{fig:NNvsFDvsAL}. This detriment in performance becomes more pronounced between $\ell = 0.4$ and $\ell = 0.2$. This change corresponds with the average geodesic distance between points in the fixed design data-set, which is equal to 0.39. This metric is shown as a dashed vertical line bottom row plot of Figure~\ref{fig:NNvsFDvsALplots}. Classifications regions with a characteristic size smaller than this value could be ignored by the classifiers, which is what we observe in the top row of Figure~\ref{fig:NNvsFDvsALplots}. Overall, we see that Gaussian process classifiers and active learning provide advantages in accuracy when compared to the baseline nearest neighbor classifier. 

\subsection{Characterization of inducibility regions}
\label{sec:inducibility}

\begin{figure}[h!]
\includegraphics[width=\textwidth]{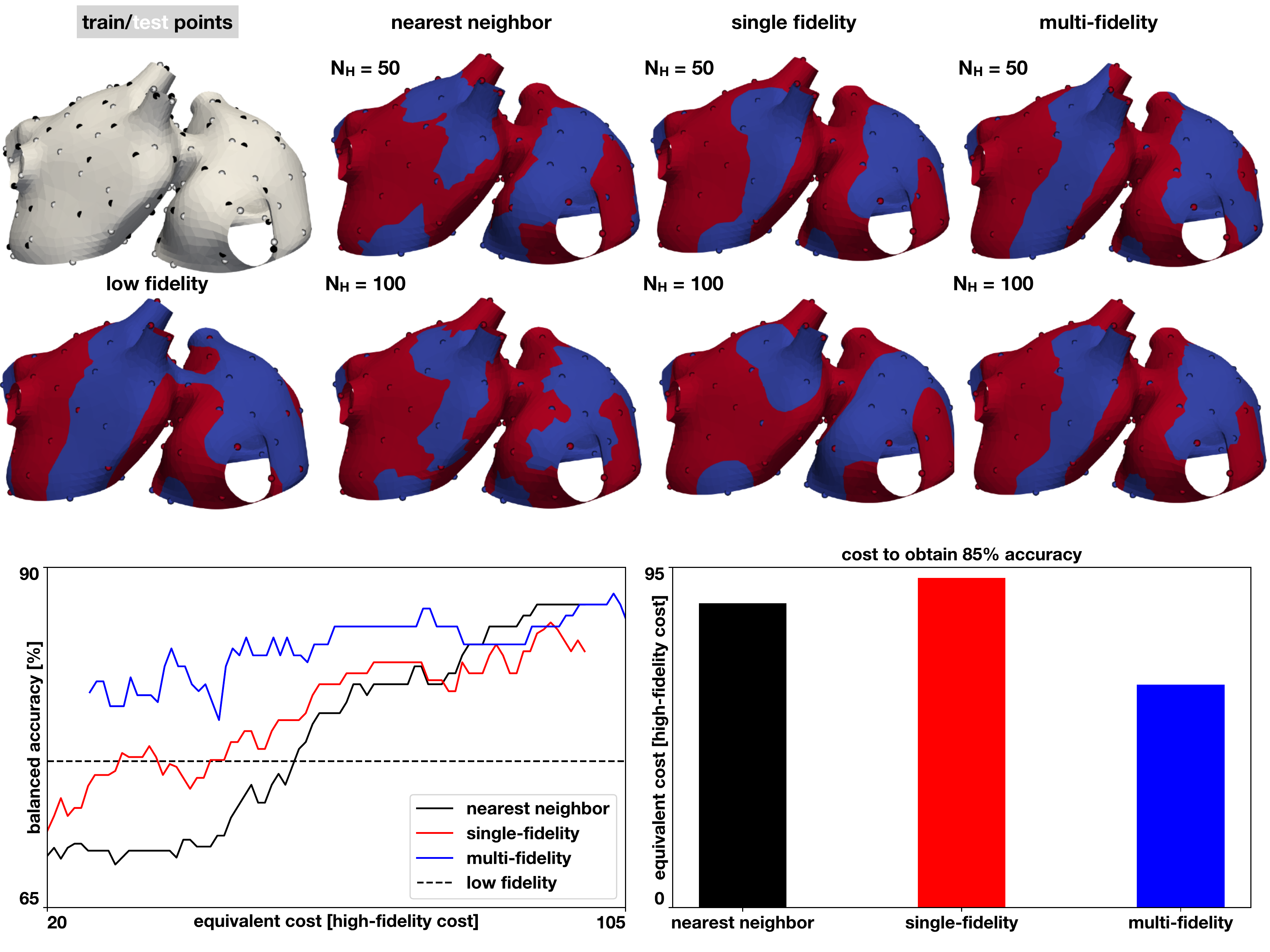}
\caption{\textbf{Inducibility maps for the baseline case.} In the upper left, we show the location of training and testing points. In the second row left, we show a single-fidelity Gaussian process classifier trained with 100 low fidelity samples. In the remainder panels, we show the nearest neighbor, single-fidelity Gaussian process classifier, and multi-fidelity Gaussian process classifier trained with 50 and 100 high fidelity samples. In the lower-left panel, we show have balanced accuracy evolves as more samples are available for the three classifiers. The samples are represented as the cost of running a high fidelity model and the multi-fidelity curve is shifted to the right to account for the cost of 100 low fidelity simulations. The dashed horizontal line represents the accuracy of a Gaussian process classifier trained with 100 low fidelity simulations predicting the high fidelity test set. The bottom right panel shows the equivalent cost in terms of high fidelity simulations to reach 85\% balanced accuracy for the three classifiers.}
\label{fig:inducibility_results}
\end{figure}

We examine the inducibility of the two models described in section~\ref{sec:modeling}: one with ablations that isolate the pulmonary veins and a baseline model, without this treatment. For each model, we create a training set and test set, both containing 100 samples, using a fixed design, as described in section~\ref{sec:numerical} and shown in Figure~\ref{fig:inducibility_results}, top left. We run the model using each of these points as a pacing site and check whether AF was induced or not. For the training set, we also run the low fidelity model, obtaining 100 samples. We test three different classifiers for both cases: a nearest neighbor classifier described in section~\ref{sec:numerical}, a single-fidelity Gaussian process classifier described in section~\ref{sec:GPclass}, and a multi-fidelity Gaussian process classifier described in section~\ref{sec:MFclass} with 100 low fidelity samples. We train the classifiers with different amounts of data from the training set, ranging from 20 to 100 points. For each level of data, we evaluate the performance of the classifier computing the balanced accuracy in the 100 samples of the test set. 

\begin{figure}[h!]
\includegraphics[width=\textwidth]{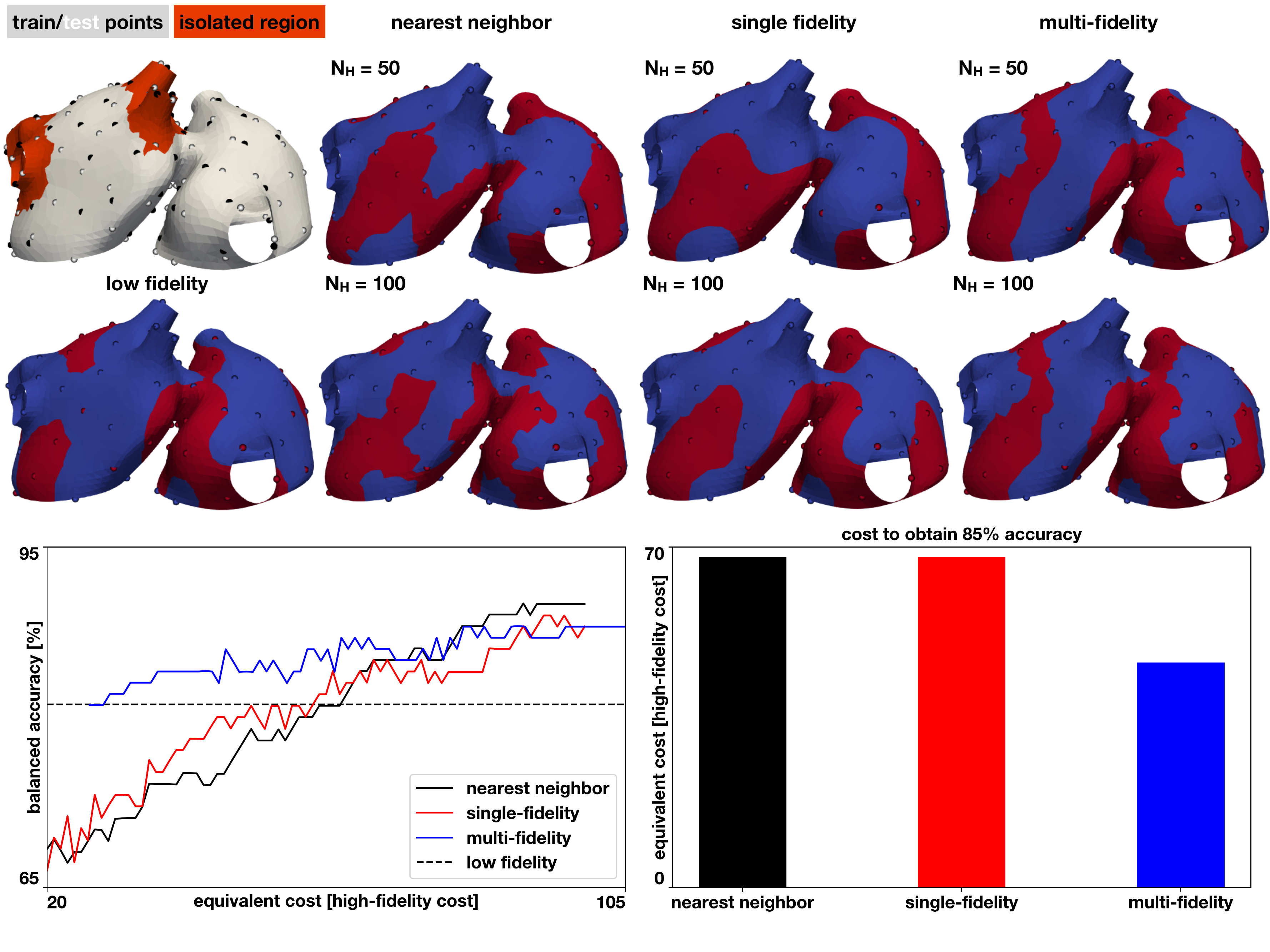}
\caption{\textbf{Inducibility maps for the pulmonary vein isolation case.} In the upper left, we show the location of training and testing points and the ablation. In the second row left, we show a single-fidelity Gaussian process classifier trained with 100 low fidelity samples. In the remainder panels, we show the nearest neighbor, single-fidelity Gaussian process classifier, and multi-fidelity Gaussian process classifier trained with 50 and 100 high fidelity samples. In the lower-left panel, we show have balanced accuracy evolves as more samples are available for the three classifiers. The samples are represented as the cost of running a high fidelity model and the multi-fidelity curve is shifted to the right to account for the cost of 100 low fidelity simulations. The dashed horizontal line represents the accuracy of a Gaussian process classifier trained with 100 low fidelity simulations predicting the high fidelity test set. The bottom right panel shows the equivalent cost in terms of high fidelity simulations to reach 85\% balanced accuracy for the three classifiers.}
\label{fig:inducibility_results_ablation}
\end{figure}

The results of this numerical experiment are summarized in Figures~\ref{fig:inducibility_results} for the untreated atria, and Figure~\ref{fig:inducibility_results_ablation} for atria treated with pulmonary vein isolation. For the baseline case, in 62\% of the samples of high fidelity training set we observed AF, while in the low fidelity training set, only 52\% of the samples presented AF. The low and high fidelity models agreed in their predictions 82\% of the time. In the high fidelity test set, we also observed that in 62\% of the cases AF was induced. In Figure~\ref{fig:inducibility_results}, top rows, we show the resulting classifiers trained with the same 50 and 100 high fidelity samples and also the low fidelity classifier trained with 100 samples. We see that the multi-fidelity classifier at 50 and 100 samples shares some features with the low fidelity classifier that are not present in the other two classifiers. Nonetheless, the multi-fidelity classifier is learning from the high fidelity data, as its balanced accuracy when trained with 50 samples is 84.4\% versus 75.7\% of the low fidelity classifier trained with 100\% samples. Both numbers are higher than the balanced accuracy of the nearest neighbor classifier, which corresponds to 73.2\%. For this model, with 50 samples, it is more accurate to use either the low fidelity classifier, the multi-fidelity classifier, or the single-fidelity classifier, which has a balanced accuracy of 78\%. These results are also reflected in the lower-left panel of Figure~\ref{fig:inducibility_results}, which shows the balanced accuracies for the different classifiers. We observe that the differences in accuracy tend to collapse as more data is available, showing small differences when 100 samples are provided to the classifiers. The multi-fidelity classifier has the biggest advantage in the small data regime, when it is trained with between 20 and 70 high fidelity samples. Surprisingly, the low fidelity classifier is more accurate than the nearest neighbor classifier up to 57 high fidelity samples. The cost of training the low fidelity classifier is approximately equivalent to the cost of acquiring 6.25 high fidelity samples, which makes the low fidelity classifier 9 times more cost-efficient than the nearest neighbor counterpart for that particular level of accuracy. Along the same line, we compare the cost to obtain an 85\% of balanced accuracy in the lower right panel of Figure~\ref{fig:inducibility_results}. In this case, the nearest neighbor classifier required 6.3\% fewer resources than the single-fidelity classifier, while the multi-fidelity classifier required 21.6\% fewer resources than the nearest neighbor classifier for this level of accuracy. 

We now present the results for the example where the pulmonary veins are isolated in Figure~\ref{fig:inducibility_results_ablation}. First, we observe that all classifiers, including the low fidelity, are able to capture the effect of the ablations around the pulmonary veins. In these regions none of the classifiers predict inducibility. For this model, 50\% of the high fidelity train set examples presented AF, while only 42\% of the low fidelity train set presented AF. The high fidelity test set presented 51\% of AF cases. These results show that the ablation treatment reduced the inducible sites by 19.3\% in the train set and 17.7\% in the test set. The low and high fidelity models agreed on 86\% of the cases. As shown in the lower-left panel of Figure~\ref{fig:inducibility_results_ablation}, the low fidelity classifier is able to predict the high fidelity test data with a balanced accuracy of 81.1\%. This value is higher than the balanced accuracy of nearest neighbor classifier up to 64 samples, as well as the single-fidelity classifier up to 60 samples. In this case, the low fidelity classifier is 10.2 times more cost-effective than the nearest neighbor classifier, and 9.6 times more cost-effective than the single-fidelity classifier, for that level of accuracy. We observe that the multi-fidelity classifier trained with 20 high fidelity samples has the same level of accuracy as the low fidelity classifier. This is expected as the high and low fidelity train sets agree in 19 of the 20 first samples. As in the baseline case, we see that the differences in accuracy are reduced as more samples are available. We also compare the cost to achieve 85\% balanced accuracy in Figure~\ref{fig:inducibility_results_ablation}, lower right panel. First, we observe that all classifiers require fewer samples to obtain this accuracy than in the baseline AF model. We see that both the nearest neighbor and the single fidelity classifier require 68 samples to obtain 85\% balanced accuracy. In contrast, the multi-fidelity classifier requires 24.7\% fewer resources to achieve this accuracy.  

Overall, in both cases, the multi-fidelity approach provided advantages over the single-fidelity classifiers, especially when a small number of samples are available.
% ablation
% train inducible 50
% test inducible 51
% LF inducible 42
% low to high accuracy 0.86
% low to test accuracy 0.7576400679117148

\section{Discussion}
\label{sec:discussion}

In this study, we propose a novel methodology to estimate the AF inducibility
regions of a computational model of the human atria.  This is
achieved by training a Gaussian process classifier that indicates
whether a given point on the atria is associated with a sustained
AF event, when incrementally pacing from its location. Our classifier is
directly trained on the atrial surface, hence it embodies the geometrical
and topological properties of the atria, which are known to be key determinants
in AF.  Gaussian process regression on Riemannian manifolds is not a novel concept,
as well as its link to certain types of SPDEs~\cite{lindgren2011}.
To the best of our knowledge, however, this is the first study proposing
a multi-fidelity Gaussian process classifier on manifolds, which extends
our previous work on Euclidean spaces~\cite{sahli2019MF}. 
The proposed method is non-intrusive, in the sense that the atrial model is 
a black-box, and outperforms classifiers based on nearest neighbor interpolation.
Moreover, when a low-fidelity model is available---in our case, obtained
by coarsening the computational mesh---, the accuracy of the classifier
can be sensibly improved with a multi-fidelity approach.
Finally, given its structure, the methodology can be easily extended to
multi-class classifier, e.g., with the capability to distinguish AF episodes
from atrial flutter. 

From a methodological perspective, our results show that the accuracy
of the classifier depends on the length scale of the inducibility region.
Intuitively, the shorter is the length scale, the more training data is needed. When the length scale is much smaller
than the size of the atria, it is more likely to observe an inducibility
region composed of disconnected and relatively small components.  Moreover,
the boundary of the inducibility region becomes less smooth.
Interestingly, the length scale has, however, a limited effect on the estimate
of the overall inducibility. This is due to fact that the volume of the
inducibility region is only marginally affected by the smoothness of
its boundary and the presence of multiple disconnected regions. We attempt to estimate the length scale of the inducibility map by training a single-fidelity classifier with both the high fidelity test and train sets.
The average length scale of the resulting classifier of the baseline AF model is $\ell = \num{0.28}$.
This is smaller than the average distance between points in the training set,
which corresponds to \num{0.39}, and may explain the balanced accuracies that we obtained were only around 90\%. We also observed in the numerical assessment
that the efficiency of active learning deteriorates at smaller length scales, for $\ell$ between
\numrange{0.2}{0.4}, and we decided not to use it for predicting inducibility maps in the non-synthetic examples, as running the cardiac electrophysiology experiments is computationally expensive.  

% This is an expected result and a common problem in approximating non-smooth
% functions. Fine-scale phenomena cannot be accurately reproduced by a coarse
% but smooth representation unless the number of points is sufficiently large
% to solve the scale. In this case, low order approximations such as nearest
% neighbor are comparable to high order representation.

From a computational viewpoint, the proposed multi-fidelity classifier reports
the maximum improvements in accuracy in a typical data set of 40 pacing sites.  In general,
the low-fidelity classifier was more accurate for a small number of
samples (less than 50), while for a larger sample size the difference
between single- and multi-fidelity classifiers is less pronounced.
When comparing the model without ablation lines and with ablation,
both high- and low-fidelity models agree on the observed
reduced inducibility due to ablation. In the case of ablation, therefore,
it is convenient to adopt a multi-fidelity approach or even just the low-fidelity
classifier, to save computational time.
In fact, the biggest advantage of the low-fidelity classifier relies
on its very limited computational cost, which is only a small fraction of the high-fidelity counterpart.  This highlights the importance of taking
advantage of these inexpensive approximations of the high-fidelity
model whenever possible. We remark that our low-fidelity model does
not require a training phase itself, thus there is no additional
offline cost.

% FSC: discussion of the results. Ablation reduced the number of inducible sites in the high and low fidelity models. Since both models agree on the effects of the ablation, it is easier for the low fidelity model to predict the high fidelity. In the cases with ablation, it makes more sense to take a multi-fidelity approach. 
    
%FSC: discussion of the results. In both cases, with and without ablation, the low fidelity classifier was more accurate in the range from 0 to 50 samples, taking only a fraction of the computational cost. This highlights the importance of taking advantage of this inexpensive approximations. 

%FSC: for the typical 40 samples reported in other studies, the multi-fidelity classifier shows the biggest improvements in accuracy 

Finally, from a modeling perspective, our results on the inducibility of AF
are in agreement with those reported in the literature.
Firstly, points in the proximity of fibrotic regions
are more likely to induce AF~\cite{Kawai2019}. Visually, there is a spatial correlation between
the inducibility region (see Figure~\ref{fig:inducibility_results} and \ref{fig:inducibility_results_ablation})
and the fibrosis distribution (Figure~\ref{fig:anatomy}).
%\textcolor{red}{FIXME: FSC, I think there is word missing here.}
The local inducibility property may therefore reflect the local tissue
properties~\cite{Boyle2021}.  Nonetheless, inducibility may also depend on other factors,
such as an abrupt change in the fiber direction, heterogeneity in the ionic parameters,
and the presence of anatomical defects or a scar. Hence, pacing sites leading
to AF may not necessarily be correlated with the local tissue properties.
Secondly, our results show that, with a fixed design, 40 pacing points are sufficient
to achieve a good estimate of the inducibility~\cite{Boyle2021}, while 20 are probably too few.
The multi-fidelity classifier, however, can achieve high accuracy
with only 20 samples.
Thirdly, the ablation treatment reduced the overall inducibility, essentially
because a large inducible region surrounding the pulmonary veins has been isolated
from the rest of the tissue, impeding the emergence of AF.  Due to the presence
of severe fibrosis in the tissue, however, it is still possible to induce AF
from several other portions of the atria, mostly unaffected by ablation.
Finally, as described above, the inducibility region in both cases shows
a small length scale, which can explain why is that pacing from different but sufficiently
close points may lead to discordant results in AF inducibility. In other
words, the uncertainty in the outcome is potentially large for some pacing sites.

Our work also presents some limitations. For instance, we limited our analysis to a single anatomy and a single fibrosis
case, but we tested an ablation treatment. Methodologically, there is no
additional difficulty in applying the approach to other cases, as the framework
is sufficiently general.  We also tested a single pacing protocol with a fixed
design.  The stimulation protocol is typically tailored to the ionic model
and can be tested in a single-cell preparation.
The duration of each simulation, \SI{4}{\s}, is sufficiently long to detect
AF events, but it might preclude the discovery of self-terminating episodes
of AF, or the translation of an AF event to atrial flutter. These cases are
typically very limited in number. The presence of self-terminating AF also
depends on the ionic model used, which may not be suitable for long
simulations (more than \SI{1}{\minute}).
Finally, we observed that using active learning can be effective in judiciously selecting new observation sites, albeit with a deteriorating efficiency at smaller length scales. Nonetheless, this limitation motivates future work on exploring new kernel functions and active
learning criteria that might be better suited for this task. 

% \subsection{Clinical relevance}
% \label{sec:clinical}

From a clinical perspective, there is an increasing application of patient-specific electrophysiology models. Thus, there is an compelling need for reducing the overall
time needed to deliver the optimal virtual treatment within the constraints dictated by clinical practice~\cite{Boyle2021,azzolin2021,pagani2021enabling}.
This study shows that
the proposed Gaussian process classifier can in fact reduce the computational
cost while maintaining a comparable or even better accuracy to a single-fidelity approach.Moreover, it does not require intrusive changes to existing
implementations and it has a very limited computational overhead, rendering its translation to existing patient-specific solutions feasible and appealing.

Inducibility maps can also offer a novel, yet unexplored, view into AF,
possibly unveiling regions susceptible to trigger AF. They
could be used to design and test ablation scenarios, e.g., by isolating
vulnerable regions. These maps could also be used to validate an AF model, by checking whether the patient-specific model and the real atria agree on the inducibility observed during a procedure.

In summary, our multi-fidelity classifier provides an efficient methodology to evaluate the effect of ablation therapy in patient-specific models of AF. We envision that this tool will accelerate the personalization of accurate treatments in the clinical setting.

% optimal distribution of pacing sites.

% FSC: Question: can we use the classifier to make optimal ablation predictions?

% \subsection{Limitations}
% \label{sec:limitations}

%%%%%%%%%%%%%%%%%%%%%%%%%%%%%%%%%%%%%%%%%%%%%%%%%%%%%%%%%%%%%%%%%%%
\section*{Acknowledgments}
This work was supported by the ANID Millennium Science Initiative Program Grant NCN17-129 and ANID-FONDECYT Postdoctoral Fellowship 3190355 to F.S.C, the DOE grant DE-SC0019116, AFOSR grant FA9550-20-1-0060, and DOE-ARPA grant DE-AR0001201 awarded to P.P. This work was also financially supported by the Theo Rossi di Montelera Foundation, the Metis Foundation Sergio Mantegazza, the Fidinam Foundation, the Horten Foundation and the CSCS--Swiss National Supercomputing Centre production grant s1074 to S.P..

\section*{Author contributions}
FSC and SP conceived the problem. FSC and PP formulated and implemented the classifier, whereas LG implemented all necessary
steps to perform multi-fidelity simulations on the supercomputer.  AG and SP provided the atrial model, the fibrosis
pattern and the ablation lines. RK and SP supervised the work of LG.
FSC, SP and LG wrote the manuscript, and all authors reviewed and improved it.

\section*{Conflict of interest}
None.
%%%%%%%%%%%%%%%%%%%%%%%%%%%%%%%%%%%%%%%%%%%%%%%%%%%%%%%%%%%%%%%%%%%

%%%%%%%%%%%%%%%%%%%%%%%%%%%%%%%%%%%%%%%%%%%%%%%%%%%%%%%%%%%%%%%%%%%

%%%%%%%%%%%%%%%%%%%%%%%%%%%%%%%%%%%%%%%%%%%%%%%%%%%%%%%%%%%%%%%%%%%
%%%%%%%%%%%%%%%%%%%%%%%%%%%%%%%%%%%%%%%%%%%%%%%%%%%%%%%%%%%%%%%%%%%

\bibliographystyle{frontiersinHLTH_FPHY} % for Health, Physics and Mathematics articles
\bibliography{main}

\end{document}